\documentclass[11pt]{article}
\usepackage{acl2015}
\usepackage{times}
\usepackage{url}
\usepackage{latexsym}
\usepackage{enumitem}
\setlist{nosep} 
\usepackage{CJKutf8}
\usepackage{graphicx}
\usepackage{color}
\usepackage{comment}

\title{Multi-Source Neural Translation}

\author{Barret Zoph and Kevin Knight  \\
Information Sciences Institute \\
Department of Computer Science \\
University of Southern California \\
{\tt \{zoph,knight\}@isi.edu}}

\date{}

\begin{document}
\maketitle

\begin{abstract}
We build a multi-source machine translation model and train it to maximize the probability of a target English string given French and German sources.  Using the neural encoder-decoder framework, we explore several combination methods and report up to +4.8 Bleu increases on top of a very strong attention-based neural translation model.
\end{abstract}

\section{Introduction}

\newcite{kay2000} points out that if a document is translated once, it is likely to be translated again and again into other languages.  This gives rise to an interesting idea: a human does the first translation by hand, then turns the rest over to machine translation (MT).   The translation system now has two strings as input, which can reduce ambiguity via ``triangulation'' (Kay's term).  For example, the normally ambiguous English word ``bank'' may be more easily translated into French in the presence of a second, German input string containing the word ``Flussufer'' (river bank).

\newcite{och01} describe such a {\em multi-source} MT system.  They first train separate bilingual MT systems $F$$\rightarrow$$E$, $G$$\rightarrow$$E$, etc.
At runtime, they separately translate input strings $f$ and $g$ into candidate target strings $e_1$ and $e_2$, then select the best one of the two.  A typical selection factor is the product of the system scores.  \newcite{lane08} revisits such factors in the context of log-linear models and Bleu score, while \newcite{MAX10.823} re-rank $F$$\rightarrow$$E$ n-best lists using n-gram precision with respect to $G$$\rightarrow$$E$ translations.  \newcite{ccb02} exploits hypothesis selection in multi-source MT to expand available corpora, via co-training.

Others use system combination techniques to merge hypotheses at the word level, creating the ability to synthesize new translations outside those proposed by the single-source translators.  These methods include confusion networks \cite{Matusov_computingconsensus,Schroeder:2009:WLM:1609067.1609147}, source-side string combination \cite{Schroeder:2009:WLM:1609067.1609147}, and median strings \cite{conf/icpr/Gonzalez-RubioC10}.

The above work all relies on base MT systems trained on {\em bilingual data}, using traditional methods.  This follows early work in sentence alignment \cite{gale1993program} and word alignment \cite{simard1999text}, which exploited trilingual text, but did not build trilingual models.  Previous authors possibly considered a three-dimensional translation table t($e|f,g$) to be prohibitive.

In this paper, by contrast, we train a P($e|f,g$) model directly on  {\em trilingual data}, and we use that model to decode an ($f,g$) pair simultaneously.  We view this as a kind of multi-tape transduction \cite{elgot1965relations,kaplan1994regular,derimake} with two input tapes and one output tape.  Our contributions are as follows:

\begin{itemize}
\item We train a P($e|f,g$) model directly on trilingual data, and we use it to decode a new source string pair ($f,g$) into target string $e$.
\item We show positive Bleu improvements over strong single-source baselines.
\item We show that improvements are best when the two source languages are more distant from each other.
\end{itemize}

We are able to achieve these results using the framework of neural encoder-decoder models, where multi-target MT \cite{dong15} and multi-source, cross-modal mappings have been explored \cite{luong15}.

\section{Multi-Source Neural MT}
 
\begin{figure*}
\begin{center}
\includegraphics[width=12cm]{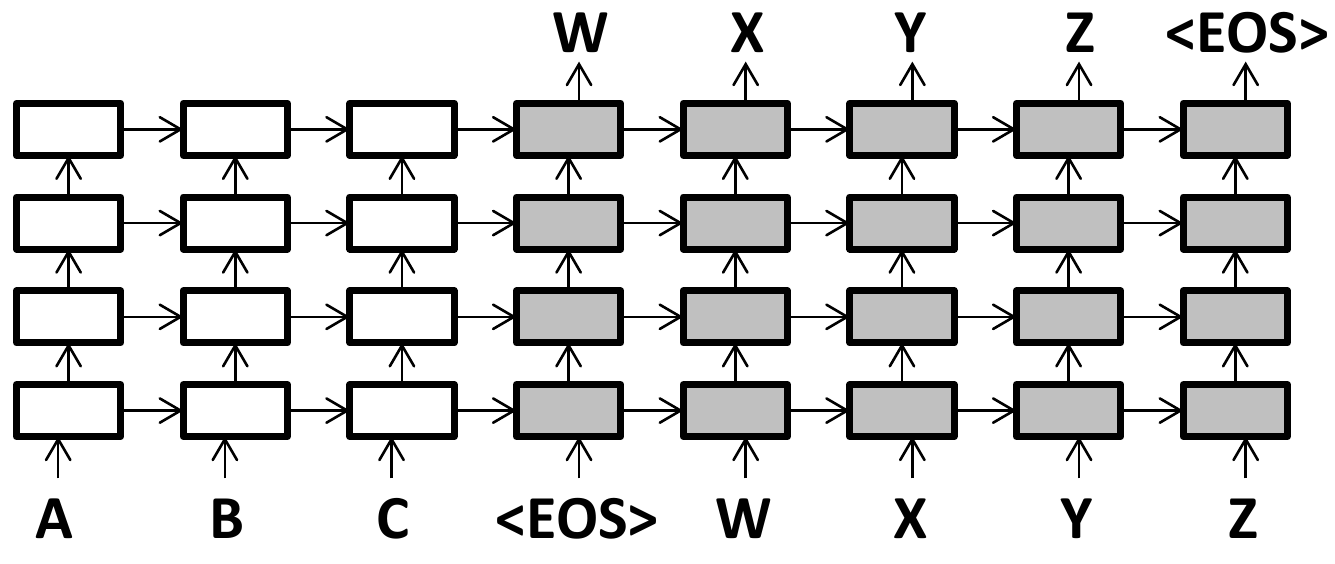}
\end{center}
\vspace{-0.1in}
\caption{The encoder-decoder framework for neural machine translation (NMT) \cite{sutskever2014sequence}.  Here, a source sentence C B A (presented in reverse order as A B C) is translated into a target sentence W X Y Z.  At each step, an evolving real-valued vector summarizes the state of the encoder (white) and decoder (gray).}
\label{enc-dec}
\end{figure*}

In the neural encoder-decoder framework for MT \cite{neco1997asynchronous,castano1997connectionist,sutskever2014sequence,bahdanau2014neural,luong2015effective}, we use a recurrent neural network ({\em encoder}) to convert a source sentence into a dense, fixed-length vector.  We then use another recurrent network ({\em decoder}) to convert that vector in a target sentence.\footnote{We follow previous authors in presenting the source sentence to the encoder in reverse order.} 

In this paper, we use a four-layer encoder-decoder system (Figure~\ref{enc-dec}) with long short-term memory (LSTM) units \cite{lstm} trained for maximum likelihood (via a softmax layer) with back-propagation through time \cite{btt}. 
For our baseline single-source MT system we use two different models, one of which implements the local attention plus feed-input model from \newcite{luong2015effective}.

\begin{figure*}
\begin{center}
\includegraphics[width=12cm]{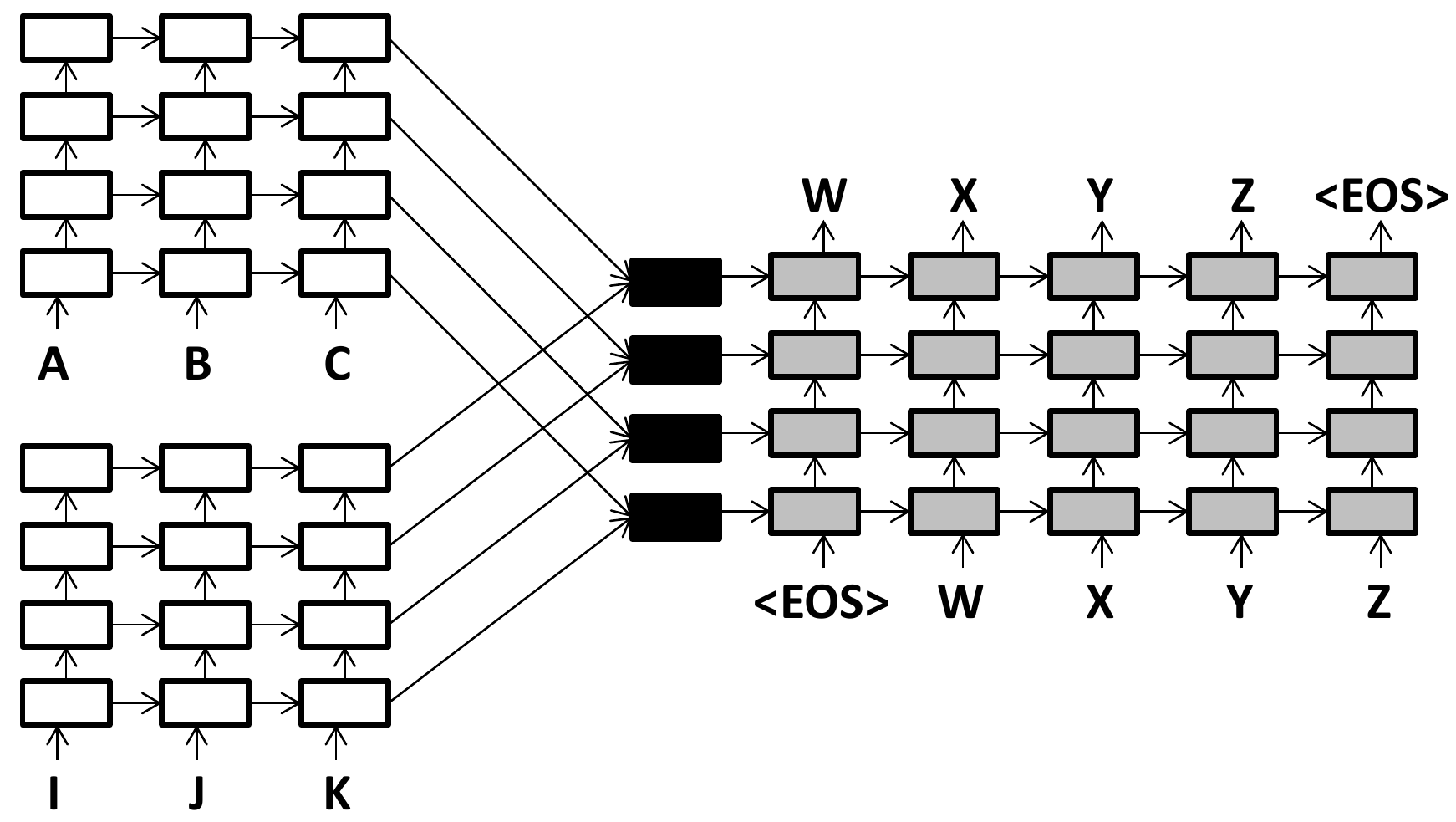}
\end{center}
\vspace{-0.1in}
\caption{Multi-source encoder-decoder model for MT\@.  We have two source sentences (C B A and K J I) in different languages. Each language has its own encoder; it passes its final hidden and cell state to a set of {\em combiners} (in black). The output of a combiner is a hidden state and cell state of the same dimension.}
\label{enc-dec-multi}
\end{figure*}

Figure~\ref{enc-dec-multi} shows our approach to multi-source MT\@. Each source language has its own encoder. The question is how to combine the hidden states and cell states from each encoder, to pass on to the decoder. Black {\em combiner} blocks implement a function whose input is two hidden states ($h_1$ and $h_2$) and two cell states ($c_1$ and $c_2$), and whose output is a single hidden state $h$ and cell state $c$. We propose two combination methods.




\subsection{Basic Combination Method}

The Basic method works by concatenating the two hidden states from the source encoders, applying a linear transformation $W_c$ (size 2000 x 1000), then sending its output through a tanh non-linearity. This operation is represented by the equation:

\begin{equation} 
h = \mathrm{tanh}\Big( W_{c}\big[ h_{1} ; h_{2} \big]  \Big) 
\end{equation}

\noindent $W_c$ and all other weights in the network are learned from example string triples drawn from a trilingual training corpus. 

The new cell state is simply the sum of the two cell states from the encoders. 

\begin{equation} 
c = c_{1} + c_{2} 
\end{equation}

\noindent
We also attempted to concatenate cell states and apply a linear transformation, but training diverges due to large cell values.

\subsection{Child-Sum Method}

Our second combination method is inspired by the  Child-Sum Tree-LSTMs of \newcite{treeLSTM}. Here, we use an LSTM variant to combine the two hidden states and cells. The standard LSTM input, output, and new cell value are all calculated. Then cell states from each encoder get their own forget gates. The final cell state and hidden state are calculated as in a normal LSTM.  More precisely:

\begin{equation} i = \mathrm{sigmoid} \Big(  W_{1}^{i}h_{1} + W_{2}^{i}h_{2}  \Big) \end{equation}
\begin{equation} f = \mathrm{sigmoid} \Big(  W_{i}^{f}h_{i} \Big) \end{equation}
\begin{equation} o = \mathrm{sigmoid} \Big(  W_{1}^{o}h_{1} + W_{2}^{o}h_{2}  \Big) \end{equation}
\begin{equation} u = \mathrm{tanh} \Big(  W_{1}^{u}h_{1} + W_{2}^{u}h_{2}  \Big) \end{equation}
\begin{equation} c = i_{f} \odot u_{f} + f_{1} \odot c_{1} + f_{2} \odot c_{2} \end{equation}
\begin{equation} h = o_{f} \odot \mathrm{tanh}(c_{f}) \end{equation}

This method employs eight new matrices (the $W$'s in the above equations), each of size 1000~x~1000. The $\odot$ symbol represents an elementwise multiplication. In equation~3, $i$ represents the input gate of a typical LSTM cell. In equation~4, there are two forget gates indexed by the subscript $i$ that serve as the forget gates for each of the incoming cells for each of the encoders. In equation~5, $o$ represents the output gate of a normal LSTM. $i$, $f$, $o$, and $u$ are all size-1000 vectors.

\subsection{Multi-Source Attention}

Our single-source attention model is modeled off the local-p attention model with feed input from \newcite{luong2015effective}, where hidden states from the top decoder layer can look back at the top hidden states from the encoder. The top decoder hidden state is combined with a weighted sum of the encoder hidden states, to make a better hidden state vector ($\tilde{h_{t}}$), which is passed to the softmax output layer. With input-feeding, the hidden state from the attention model is sent down to the bottom decoder layer at the next time step.  

The local-p attention model from \newcite{luong2015effective} works as follows. First, a position to look at in the source encoder is predicted by equation~9:

\begin{equation}
p_t = S \cdot \mathrm{sigmoid} \big( v_p^T \mathrm{tanh}\big( W_p h_t \big) \big )
\end{equation}

\noindent
$S$ is the source sentence length, and $v_p$ and $W_p$ are learned parameters, with $v_p$ being a vector of dimension 1000, and $W_p$ being a matrix of dimension 1000~x~1000.

After $p_t$ is computed, a window of size $2D+1$ is looked at in the top layer of the source encoder centered around $p_t$ ($D=10$). 
For each hidden state in this window, we compute an alignment score $a_t\big(s\big)$, between 0 and 1. This alignment score is computed by equations~10, 11 and 12:

\begin{equation}
a_t\big(s\big) = \mathrm{align}\big( h_t , h_s \big) \mathrm{exp} \Big( \frac{-\big( s - p_t \big)^2}{2\sigma^2} \Big)
\end{equation}

\begin{equation}
\mathrm{align}\big( h_t , h_s \big) = \frac{\mathrm{exp}\big(\mathrm{score}\big( h_t , h_s \big) \big) }{\sum_{s'}\mathrm{exp}\big(\mathrm{score}\big( h_t , h_{s'} \big) \big)}
\end{equation}

\begin{equation}
\mathrm{score}\big( h_t , h_s \big) = h_t^T W_a h_s
\end{equation}

In equation~10, $\sigma$ is set to be $D/2$ and $s$ is the source index for that hidden state. $W_a$ is a learnable parameter of dimension 1000~x~1000. 

Once all of the alignments are calculated, $c_t$ is created by taking a weighted sum of all source hidden states multiplied by their alignment weight. 

The final hidden state sent to the softmax layer is given by:

\begin{equation} 
\tilde{h_{t}} = tanh \Big( W_{c} \big[ h_{t} ; c_{t} \big] \Big) 
\end{equation}


We modify this attention model to look at both source encoders simultaneously. We create a context vector from each source encoder named $c_{t}^{1}$ and $c_{t}^{2}$ instead of the just $c_t$ in the single-source attention model:

\begin{equation}
\tilde{h_{t}} = tanh \Big( W_{c} \big[ h_{t} ; c_{t}^{1} ; c_{t}^{2} \big] \Big)
\end{equation}


\begin{figure}
\begin{tabular}{|l|r|r|r|} \hline
& French & English & German\\ \hline
Word tokens & 66.2m & 59.4m & 57.0m \\ \hline
Word types & 424,832 & 381,062 & 865,806 \\ \hline
Segment pairs & \multicolumn{3}{c|}{2,378,112} \\ \hline
Ave. segment & 27.8 & 25.0 & 24.0 \\ 
length (tokens) & & & \\ \hline
\end{tabular}
\caption{Trilingual corpus statistics.}
\label{corpus-stats}
\end{figure}

In our multi-source attention model we now have two $p_t$ variables, one for each source encoder. We also have two separate sets of alignments and therefore now have two $c_t$ values denoted by $c_t^1$ and $c_t^2$ as mentioned above. We also have distinct $W_a$, $v_p$, and $W_p$ parameters for each encoder.

\section{Experiments}

We use English, French, and German data from a subset of the WMT 2014 dataset \cite{wmt2014}.  Figure~\ref{corpus-stats} shows statistics for our training set. For development, we use the 3000 sentences supplied by WMT. For testing, we use a 1503-line trilingual subset of the WMT test set.


For the single-source models, we follow the training procedure used in \newcite{luong2015effective}, but with 15 epochs and halving the learning rate every full epoch after the 10th epoch.  We also re-scale the normalized gradient when norm~$>$~5. For training, we use a minibatch size of 128, a hidden state size of 1000, and dropout as in \newcite{dropout}. The dropout rate is 0.2, the initial parameter range is \mbox{$[$-0.1,~+0.1$]$}, and the learning rate is 1.0. For the normal and multi-source attention models, we adjust these parameters to 0.3, \mbox{$[$-0.08,~+0.08$]$}, and 0.7, respectively, to adjust for overfitting.

\begin{figure}
\begin{tabular}{|l|l|r|r|} \hline
\multicolumn{4}{|c|}{Target = English} \\ \hline
{\bf Source} & {\bf Method} & {\bf Ppl} & {\bf BLEU} \\ \hline
French & --- & 10.3 & 21.0 \\ \hline
German & --- &  15.9 & 17.3 \\ \hline
French+German &  Basic & 8.7 & 23.2 \\ \hline \hline
French+German &  Child-Sum & 9.0 & 22.5 \\ \hline 
French+French & Child-Sum & 10.9 & 20.7 \\ \hline \hline
French & Attention & 8.1 & 25.2 \\ \hline
French+German & B-Attent. & 5.7 & 30.0 \\ \hline
French+German & CS-Attent. & 6.0 & 29.6\\ \hline
\end{tabular}
\caption{Multi-source MT for target English, with source languages French and German.  Ppl reports test-set perplexity  as the system predicts English tokens.  BLEU is scored using the multi-bleu.perl script from Moses. For our evaluation we use a single reference and they are case sensitive.}
\label{results-english}
\end{figure}

Figure~\ref{results-english} show our results for target English, with source languages being French and German).
We see that the Basic combination method yields a +4.8 Bleu improvement over the strongest single-source, attention-based system.  It also improves Bleu by +2.2 over the non-attention baseline.  The Child-Sum method gives improvements of +4.4 and +1.4. We also confirm that two copies of the same French input yields no BLEU improvement.  

\begin{figure}
\begin{center}
\includegraphics[width=7.7cm]{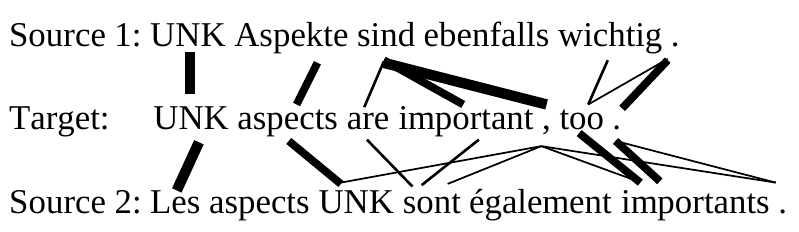}
\end{center}
\vspace{-0.2in}
\caption{Action of the multi-attention model as the neural decoder generates target English from French/German sources (test set).  Lines show strengths of $a_t(s)$.}
\label{align}
\end{figure}

Figure~\ref{align} shows the action of the multi-attention model during decoding.

When our source languages are English and French (Figure~\ref{results-german}), we observe smaller BLEU gains (up to +1.1).  This is evidence that the more distinct the source languages, the better they disambiguate each other.

\begin{figure}
\begin{tabular}{|l|l|r|r|} \hline
\multicolumn{4}{|c|}{Target = German} \\ \hline
{\bf Source} & {\bf Method} & {\bf Ppl} & {\bf BLEU} \\ \hline
French & --- & 12.3 & 10.6 \\ \hline
English & --- & 9.6 & 13.4 \\ \hline \hline
French+English &  Basic & 9.1 & 14.5 \\ \hline 
French+English &  Child-Sum & 9.5 & 14.4 \\ \hline \hline 
 English & Attention & 7.3 & 17.6 \\ \hline
 French+English & B-Attent. & 6.9 & 18.6 \\ \hline
 French+English & CS-Attent. & 7.1 & 18.2\\ \hline
\end{tabular}
\caption{Multi-source MT results for target German, with source languages French and English.}
\label{results-german}
\end{figure}

\section{Conclusion}

We describe a multi-source neural MT system that gets up to +4.8 Bleu gains over a very strong attention-based, single-source baseline.  We obtain this result through a novel encoder-vector combination method and a novel multi-attention system. We release the code for these experiments at \url{https://github.com/isi-nlp/Zoph_RNN}.

%
%
%

\bibliographystyle{acl}
\bibliography{multisource}

\end{document}